# ¿Es Posible Aplicar las Redes Neuronales Artificiales en el Pronóstico Sísmico? Análisis Preliminar Aplicando la Topología Radial. Caso: México

# Can Artificial Neural Networks be Applied in Seismic Predicition? Preliminary Analysis Applying Radial Topology. Case: Mexico

Cinthya Mota-Hernández[1], Luis Esquivel-Rodríguez[1] and Rafael Alvarado-Corona[2]

[1]Universidad Autónoma del Estado de México, México. www.uaemex.mx

[2]SEP, SARACS Research Group, www.saracs.com.mx, Ciudad de México, México. E-mail: ralvcor@gmail.com

**Resumen---** Sismos tectónicos de elevada magnitud pueden causar pérdidas considerables en términos de vidas humanas, económicos y de infraestructura, entre otros. De acuerdo con una evaluación publicada por el Servicio Geológico de los Estados Unidos, 30 es el número de sismos que han impactado en gran medida a México desde finales del siglo XIX al actual.. En base a datos del Servicio Sismológico Nacional, en el periodo comprendido entre el 1 de Enero del 2006 y el 1 de Mayo del 2013 se han presentado 5,826 sismos mayores a 4.0 grados de magnitud en la escala de Richter (28.54% del total de sismos registrados en el territorio nacional), siendo la placa del Pacífico y la placa de Cocos las de mayor importancia. La investigación describe el desarrollo de una Red Neuronal Artificial (RNA) basada en la topología radial que busca generar una predicción con un margen de error menor al 20% que informe sobre la probabilidad de un sismo futuro, una de las preguntas principales del enfoque propuesto es: ¿Es posible aplicar las redes neuronales artificiales en el pronóstico sísmico? Se puede argumentar que la investigación tiene el potencial de aportar en el pronóstico sísmico, más investigación es necesaria para consolidar datos y contribuir a mitigar el impacto causado por tales eventos al vincularse con la sociedad.

*Palabras clave--- Análisis, México, Pronóstico, Redes neuronales artificiales, Sismicidad.*

**Abstract---** Tectonic earthquakes of high magnitude can cause considerable losses in terms of human lives, economic and infrastructure, among others. According to an evaluation published by the U.S. Geological Survey, 30 is the number of earthquakes which have greatly impacted Mexico from the end of the XIX century to this one. Based upon data from the National Seismological Service, on the period between January 1, 2006 and May 1, 2013 there have occurred 5,826 earthquakes which magnitude has been greater than 4.0 degrees on the Richter magnitude scale (25.54% of the total of earthquakes registered on the national territory), being the Pacific Plate and the Cocos Plate the most important ones. This document describes the development of an Artificial Neural Network (ANN) based on the radial topology which seeks to generate a prediction with an error margin lower than 20% which can inform about the probability of a future earthquake one of the main questions is: can artificial neural networks be applied in seismic forecasting? It can be argued that research has the potential to bring in the forecast seismic, more research is needed to consolidate data and help mitigate the impact caused by such events linked with society.

## I. INTRODUCCION

Los desastres naturales pueden causar alto impacto en sociedades vulnerables, existe investigación que se ha realizado en varias facetas [9], por otro lado, para el monitoreo y alerta de actividad sísmica en México se cuenta a partir de 1991 con el Sistema de Alerta Sísmica (SAS) [1], que cubre las zonas aledañas a Guerrero y cuenta con un total de 12 estaciones sensoras, además del Sistema de Alerta Sísmica de la Ciudad de Oaxaca (SASO), desarrollado en el 2003 y con un total de 36 estaciones sensoras. Estos sistemas han emitido 78 alertas tempranas de más de 2350 sismos [2], hecho que ha permitido contemplar la posibilidad de extender la cobertura de los mismos a los estados comprendidos entre Jalisco y Chiapas (la mayor parte del territorio se ubica sobre la placa de Cocos). De cualquier manera, ambos sistemas reaccionan en cuanto la actividad sísmica es detectada, sin embargo, se evaluó la posibilidad de emplear una RNA capaz de buscar un patrón en los registros de sismos desde los 4.0 grados de magnitud a partir del año 1998 hasta el 1 de Mayo del 2013, la RNA se basó en una topología radial. Haciendo uso de una muestra de poco más de 5,000 sismos, misma que fue dividida para integrar los conjuntos de validación y producción, se llevó a cabo el entrenamiento de la RNA mediante el aprendizaje supervisado con el objetivo de obtener a sus salidas la magnitud y otros datos de un posible sismo futuro. El presente artículo explica el desarrollo de dicha red con base en una metodología de análisis para sistemas suaves [3], además de los resultados obtenidos del entrenamiento. Existen diversas investigaciones en el área sísmica con enfoques que enriquecen y aportan de manera integral al análisis del fenómeno [10].

## II. RECOPILACIÓN Y TRATAMIENTO DE DATOS

El primer proceso llevado a cabo fue la adquisición y procesamiento de los datos de tal manera que el entrenamiento contara con un número aceptable de muestras, cada una de ellas completa y en una escala coherente general. Los datos fueron extraídos del Servicio Sismológico Nacional [4] y se sitúan entre el 2 de Marzo del año 2006 y el 1 de Mayo del año 2013, integrando una muestra de un total de 5,798 registros sísmicos, población que fue segmentada de la siguiente forma:

TABLA I
Segmentación de datos

|  | Conjunto de entrenamiento | Conjunto de validación | Conjunto de producción |
|---|---|---|---|
| No. de muestras | 3001 | 1551 | 1246 |
| % Total | 51.76% | 26.75% | 21.49% |

Tras la formación de los grupos se llevó a cabo un análisis sobre la importancia para el caso de los campos que conformaban las muestras extraídas, optando por el empleo de las siguientes variables:

TABLA II
Variables de entrada y salida

| Entradas | Salidas |
|---|---|
| Fecha | Latitud |
| Hora | Longitud |
| Minuto | Profundidad |
| Zona | Magnitud |

El proceso de conversión de datos tanto en tipo como en escala se llevó a cabo en todos los campos que representarían las variables de entrada para la red, y en todos los conjuntos obtenidos en el paso anterior.

## III. DESARROLLO DE LA RED NEURONAL ARTIFICIAL

Partiendo del análisis del problema y con base en el planteamiento de las variables analizadas, se procedió a realizar el modelado de las medidas de desempeño de la red a entrenar, así como el análisis funcional de la misma. Para el entrenamiento se planteó el uso de aprendizaje supervisado, con una terminación basada en el error, para propósitos de exactitud se establece que el error fuera causa de terminación del entrenamiento cuando se alcanzara la condición de sobreentrenamiento; el error máximo planteado para las salidas de la red fue del 20%, cualquier salida por encima de ese rango se consideraría como razón para clasificar la red como no óptima para su uso.

El diseño elegido para estructura de la red, considerando el tratamiento de datos para la investigación, se ilustra en la Fig.1.

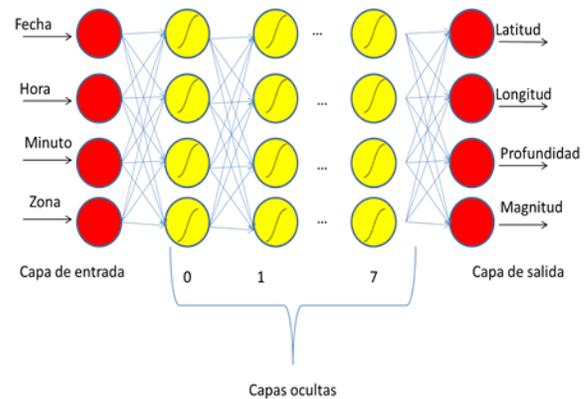

Fig. 1 Estructura de la red
Fuente: Elaboración propia

A priori, era deseable establecer el modelo de red que fuera más óptimo para la solución del problema planteado, tanto en el tiempo requerido como en el error mínimo obtenido a la salida de la red. Este proceso se basó en los aspectos siguientes:

*A. Modelos de Redes elegidos.*

1. Perceptrón multicapa, modelo conocido por ser apropiado para una gran cantidad de problemas [5].
2. Recurrente en series de tiempo [6].
3. Recurrente generalizada, similar a la anterior y comúnmente usada en predicción [7].
4. Red de base radial con perceptrón multicapa [8].
5. Radial, de propósito general.

*B. Número de elementos de procesamiento (EP).*

Las redes preliminares entrenadas poseían 7 capas ocultas, cuyas cantidades de EP son descritos en la Tabla III.

TABLA III
EP por capa

|  | No. de EP |
|---|---|
| Capa 1 | 20 |
| Capa 2 | 60 |
| Capa 3 | 100 |
| Capa 4 | 150 |
| Capa 5 | 100 |
| Capa 6 | 60 |
| Capa 7 | 20 |

*C. Total de Muestras a usar.*

En todas las redes preliminares se hizo uso de una población de 200 muestras para el entrenamiento.

*D. Reglas de Propagación y Funciones de Activación.*

Se hizo uso de las reglas de propagación momentum para el perceptrón multicapa entrenado (ésta regla es propia de los perceptrones multicapa) y Quickprop (que intenta hacer uso de un modelo cuadrático de la superficie de error para acelerar la convergencia).

*E. Criterio de convergencia.*

Se estableció como criterio de terminación el mismo a utilizar en la red final, es decir, terminación basada en el error mínimo antes de alcanzar el sobreentrenamiento. Se estableció como error mínimo el 10%. Al finalizar el entrenamiento de las redes antes mencionadas, se obtuvieron los siguientes resultados a fin de seleccionar el modelo de red más óptimo para su aplicación se llevaron a cabo las comparativas siguientes:

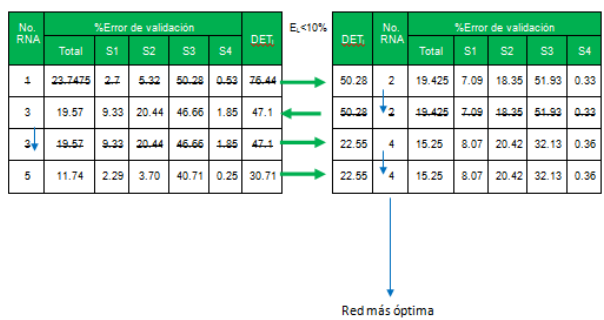

Fig. 2 Comparación de conjuntos de validación
Fuente: Elaboración propia

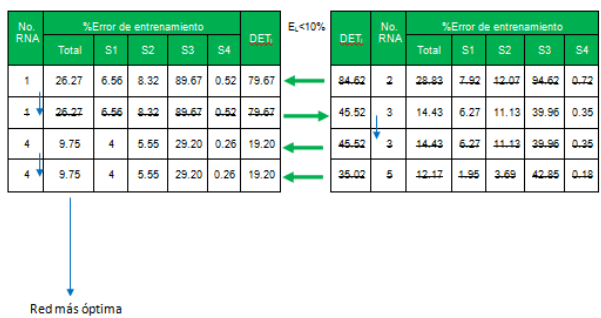

Fig. 3 Comparación de conjuntos de entrenamiento
Fuente: Elaboración propia

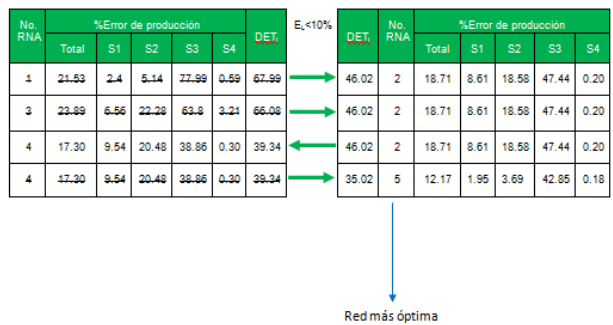

Fig. 4 Comparación de conjuntos de producción
Fuente: Elaboración propia

La comparación entre los conjuntos de entrenamiento y validación demostró que la red 4 (red de base radial con perceptrón multicapa) era la más adecuada para resolver el problema. La red final se basó por consiguiente, en una topología radial con perceptrón multicapa con los siguientes datos:

TABLA V
Detalles de la red final

|  | No. de EP | Función de transferencia | Regla de propagación |
|---|---|---|---|
| Capa de entrada | 100 | Gaussiana | Quickprop |
| Capa 1 | 50 | Tangente hiperbólica | Quickprop |
| Capa 2 | 100 | Tangente hiperbólica | Quickprop |
| Capa 3 | 200 | Tangente hiperbólica | Quickprop |
| Capa 4 | 400 | Tangente hiperbólica | Quickprop |
| Capa 5 | 200 | Tangente hiperbólica | Quickprop |
| Capa 6 | 100 | Tangente hiperbólica | Quickprop |
| Capa 7 | 50 | Tangente hiperbólica | Quickprop |

Obteniendo, a sus salidas lo siguiente:

TABLA VI
Salidas de la red final

| Dato | Red final |
|---|---|
| Ciclos requeridos | 4411 |
| Tiempo requerido | 6:24:22 |
| Conjunto de producción |  |
| Error promedio Latitud | 2.40% |
| Error promedio Longitud | 5.14% |
| Error promedio Magnitud | 0.59% |
| Conjunto de validación |  |
| Error promedio Latitud | 2.70% |
| Error promedio Longitud | 5.32% |
| Error promedio Magnitud | 0.53% |

| Conjunto de entrenamiento | |
|---|---|
| Error promedio Latitud | 6.56% |
| Error promedio Longitud | 8.32% |
| Error promedio Magnitud | 0.52% |
| MSE | 0.161018021136 |
| NMSE | 1.066836403099 |
| % Error | 12.519098354399 |

La comparación generada entre los valores actuales y los generados por la red neuronal, se muestran en la Fig. 5

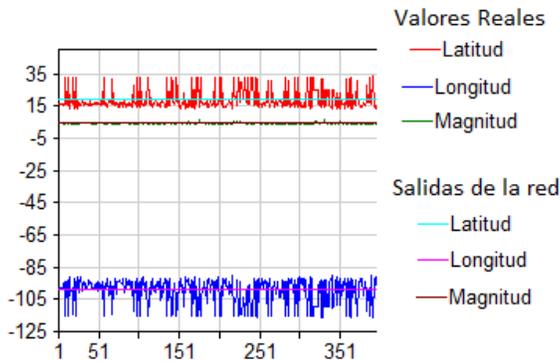

Fig. 5 Gráfica comparativa de los valores de las entradas y salidas de la red
Fuente: Elaboración propia

## IV. DISCUSIÓN

En el proceso que conlleva a la elección del modelo de RNA más adecuado para la resolución de un problema, es necesario evaluar aspectos como el tipo de problema a resolver, el error mínimo esperado a la salida, el número de muestras y el tratamiento-conversión de los datos contenidos. Pese a que la elección del número de neuronas además del número y composición de las capas ocultas es un factor limitante para ciertos equipos informáticos en cuanto a los recursos con que se cuenta, su elección correcta es clave para alcanzar el criterio de convergencia establecido. El proceso de elección del mejor modelo de RNA implica prueba y error utilizando distintos modelos y/o variaciones en su aplicación. Las redes neuronales artificiales tienen potencial aplicación en el ámbito sísmico.

## V. CONCLUSIÓNES

Un análisis de la aplicación de Redes Neuronales artificiales aplicadas al ámbito sísmico fue presentado. En conclusión, se obtuvo un error mínimo ubicado por debajo del 10% en las tres variables de salida, pese a que una de las salidas planteadas fue descartada, siendo el modelo de topología radial con perceptrón multicapa el de mayor eficiencia para la resolución del problema. La variable que en todos los casos presentó el menor error fue la magnitud sísmica; no obstante haber excluido variables de salida referentes a tiempo, se plantea llevar a cabo mayor investigación sólida que invite a la reflexión y pueda brindar resultados concluyentes.